\documentclass[lettersize,journal]{IEEEtran}
\usepackage{amsmath,amsfonts}

\usepackage{array}
\usepackage[caption=false,font=normalsize,labelfont=sf,textfont=sf]{subfig}
\usepackage{textcomp}
\usepackage{stfloats}
\usepackage{url}
\usepackage{verbatim}
\usepackage{graphicx}
\usepackage{cite}
\usepackage{booktabs}

\usepackage{algorithm}

\usepackage{algpseudocode}

\graphicspath{{figures/}}
\usepackage[colorlinks=true,linkcolor=blue,urlcolor=blue,citecolor=blue]{hyperref}
\makeatletter
\def\hlinew#1{%
  \noalign{\ifnum0=`}\fi\hrule \@height #1 \futurelet
   \reserved@a\@xhline}
\usepackage{multirow}
\usepackage{pifont}
\newcommand{\cmark}{\ding{51}}
\newcommand{\xmark}{\ding{55}}

\begin{document}

\title{CangLing-KnowFlow: A Unified Knowledge-and-Flow-fused Agent for Comprehensive Remote Sensing Applications}

\author{Zhengchao Chen, Haoran Wang, Jing Yao,~\IEEEmembership{Senior Member,~IEEE,}
        Jianshe Zhang, Pedram Ghamisi,~\IEEEmembership{Senior Member,~IEEE,}
        Jun Zhou,~\IEEEmembership{Fellow,~IEEE,}
        Peter M. Atkinson,
        Bing Zhang,~\IEEEmembership{Fellow,~IEEE}

\thanks{Z. Chen, H. Wang, J. Yao, and B. Zhang are with the State Key Laboratory of Remote Sensing and Digital Earth, Aerospace Information Research Institute, Chinese Academy of Sciences, Beijing 100094, China (e-mail: chenzc@radi.ac.cn; wanghaoran23@mails.ucas.ac.cn; jasonyao92@gmail.com; zb@radi.ac.cn).}
\thanks{J. Zhang is with the Beijing Tiandi Shijie Technology Co., Ltd., Beijing 101499, China (e-mail: zhangjianshe@gmail.com).}
\thanks{P. Ghamisi is with the Faculty of Science and Technology, Lancaster
University, Lancaster, LA1 4YQ, U.K., the Faculty of Electrical and Computer Engineering, University of Iceland, 101 Reykjavik, Iceland, and also with the
Helmholtz-Zentrum Dresden-Rossendorf, Freiberg 09599, Germany (e-mail:
p.ghamisi@gmail.com).}
\thanks{J. Zhou is with the School of Information and Communication Technology, Griffith University, Nathan, QLD 4111, Australia (e-mail: jun.zhou@griffith.edu.au).}
\thanks{P. M. Atkinson is with the Faculty of Science and Technology, Lancaster University, Lancaster, LA1 4YQ, U.K. (e-mail: pma@lancaster.ac.uk).}
}

\markboth{}%
{Shell \MakeLowercase{\textit{et al.}}: A Sample Article Using IEEEtran.cls for IEEE Journals}


\maketitle

\begin{abstract}
Automated and intelligent processing of massive remote sensing datasets is critical in Earth observation.
Existing automated systems are normally task-specific, lacking a unified framework to manage diverse, end-to-end workflows from data preprocessing to advanced interpretation across diverse RS applications. 
To fill this gap, this paper introduces CangLing-\textbf{KnowFlow}, a unified intelligent agent framework that integrates a procedural knowledge base, a dynamic workflow adjustment mechanism, and an evolutionary memory module. 
The procedural knowledge base, comprising 1,008 expert-validated workflow cases across 162 practical remote sensing tasks, guides planning and substantially reduces hallucinations common in general-purpose agents. 
During runtime failures, the dynamic workflow adjustment module autonomously diagnoses and replans recovery strategies, while the evolutionary memory module continuously learns from these events, iteratively enhancing the agent’s knowledge and performance. 
This synergy enables CangLing-KnowFlow to adapt, learn, and operate reliably across diverse and complex tasks.
We evaluated CangLing-KnowFlow on the KnowFlow-Bench, a novel benchmark of 324 workflows inspired by real-world applications, testing its performance across 13 top large language model backbones, from open-source to commercial. 
Across all complex tasks, CangLing-KnowFlow surpassed the Reflexion baseline by at least 4\% in task success rate. 
As the first most comprehensive validation along this emerging field, this research demonstrates the great potential of CangLing-KnowFlow as a robust, efficient, and scalable automated solution for complex Earth observation challenges by leveraging expert knowledge (\textbf{Know}) into adaptive and verifiable procedures (\textbf{Flow}).
\end{abstract}


\section{Introduction}
\IEEEPARstart{R}{ecent} advances in Earth observation technologies have led to a rapid growth in the availability of remote sensing (RS) data, marked by increasing volume, diversity, and temporal resolution \cite{vivone2024deep,zhang2022progress}.
While this data surge offers transformative opportunities for applications such as environmental monitoring and disaster response, it also highlights the limitations of traditional RS analysis pipelines.
These conventional methods largely dependent on manually crafted workflows designed by domain experts, therefore struggling with inefficiency, high labor demands, and limited scalability, which creates a critical bottleneck to timely and automated extraction of actionable intelligence from large and complex datasets \cite{zhang2019remotely,zhang2025core}. 

In parallel, breakthroughs in Artificial Intelligence (AI), driven by the rise of large language models (LLMs), have introduced the AI Agent as a transformative operational paradigm. AI agents possess the ability to interpret high-level tasks, plan and execute multi-step workflows, dynamically invoke specialized tools, and perform self-correction. 
These capabilities enable the seamless integration of previously disjointed, human-dependent processes into coherent and automated pipelines. This paradigm shift from human-driven methodologies to machine-autonomous operations opens a new pathway toward realizing the next generation of autonomous Earth observation systems \cite{li2025segearth}.

Building a robust and reliable RS agent is challenging due to the inherent limitations of general-purpose agent architectures. 
First, monolithic single-agent systems often produce erroneous or scientifically unsound plans, which is a phenomenon known as planning hallucination \cite{zhu2025knowagent}, especially when dealing with long-horizon tasks requiring deep domain expertise. 
Second, although traditional multi-agent systems promote collaboration, their decentralized communication protocols can introduce excessive overhead and misaligned goals.
This could create uncertainty in the process and compromise the reproducibility required for scientific analysis.
Finally, fully fixed workflows, although stable and compliant, lack the flexibility to accommodate the variability inherent in RS data (e.g., sensor types, weather conditions, temporal dynamics) and recover from unexpected runtime failures, yielding a persistent trade-off between reliability and adaptability \cite{shi2025flowagent}.

\begin{figure*}[!t]
\centering
\includegraphics[width=1\linewidth]{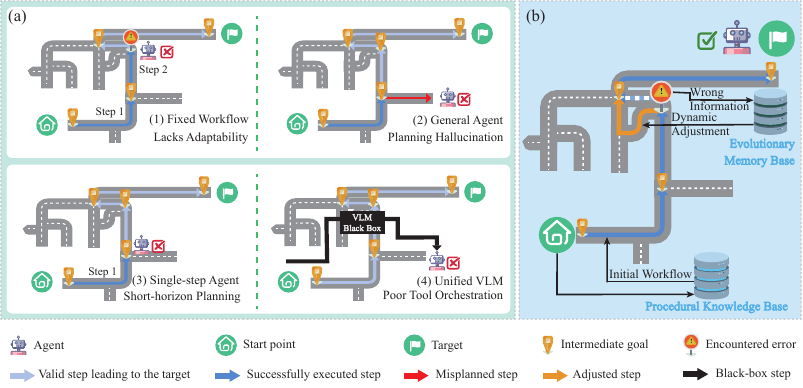} 
\caption{Conceptual comparison of different agent architectures for remote sensing. \textbf{(a) Existing agent paradigms exhibit fundamental limitations:} (1) Fixed workflows are inflexible; (2) General-purpose agents lack domain knowledge, leading to planning hallucinations; (3) Single-tool-use agents lack complex planning capabilities; and (4) Unified VLMs lack procedural transparency and robustness. \textbf{(b) Our proposed CangLing-KnowFlow framework} addresses these limitations by integrating an expert-informed knowledge base, a dynamic adjustment mechanism, and an evolutionary memory module.}
\label{fig:overall_framework}
\end{figure*}

Despite their considerable potential, the application of agent-based systems in RS remains in an exploratory phase, currently following two main directions. 
The first encompasses tool-augmented agents, including single-agent systems for tool invocation \cite{xu2024rs, wei2025geotool, chen2024llm, ning2024llm} and multi-agent frameworks tailored for specific data types \cite{lin2025shapefilegpt, lee2025multi, yu2024mineagent}. 
While these efforts validate the feasibility of agents interfacing with specialized tools, their operational scope is often confined to single-turn interactions or predefined toolsets, exhibiting limited capacity for long-term, autonomous planning and adaptation. 
The second direction is defined by unified Vision Language Models (VLMs) \cite{hu2025rsgpt, zhan2025skyeyegpt, soni2025earthdial}, attempting to handle various RS tasks end-to-end within a single massive model. 
Although these models demonstrate notable multitasking capabilities, on one hand, their monolithic nature precludes the precise orchestration of complex external toolchains, and on the other hand, their lack of workflow-level robustness renders them difficult to diagnose and correct upon failure. 
The architectural deficiencies of the existing paradigms mentioned above constitute a significant research gap, which is conceptually illustrated in Figure \ref{fig:overall_framework} (a).

More recently, agentic workflows have emerged as a solution to bridge these divergent technical paths, blancing tool flexibility with the strict requirements of scientific workflows \cite{li2025designing,feng2025earth}. 
By orchestrating agents through structured processes, this paradigm ensures analytical transparency while retaining the ability to dynamically compose toolchains.
Although early geospatial works have explored this direction \cite{bhattaram2025geoflow}, it still suffers from notable limitations.
First, they rely excessively on the generalizaion capabilities of LLMs for planning. 
Although LLMs possess extensive general knowledge, their outputs, optimized for universality, often lack the requisite specificity and depth for specialized scientific domains. 
An operational agent must be infused with the tacit knowledge, operational heuristics, and industry standards of domain experts, a capability underdeveloped in existing workflows. 
Second, they predominantly follow a static plan-once, execute-sequentially model, making them fragile when faced with runtime failures caused by data or environmental factors. 
Third, current workflow agents typically operate without a persistent memory mechanism and fail to learn from past successes or failures as some generalized agents \cite{shinn2023reflexion}, limiting their potential for continuous intelligence growth.

To systematically address the challenges mentioned above, this paper introduces a novel RS agent framework and a corresponding evaluation benchmark. 
The framework, centered on a dynamic and evolutionary agentic workflow, is designed not only to execute tasks efficiently, but also to evolve into a continuously improving domain expert. Our core contributions are fourfold:

\begin{itemize}
 \item \textbf{Codified expert knowledge facilitates scientifically valid planning.} We contribute a procedural knowledge base where expert domain knowledge is codified into a library of workflow templates. This provides a robust basis for the agent's planning, fundamentally constraining LLM-induced fallacies and ensuring the scientific validity of generated plans.

 \item \textbf{Autonomous real-time adaptation ensures robust performance in dynamic conditions.} Our framework empowers the agent with real-time monitoring and dynamic adjustment capabilities. When a procedural step fails, the agent can autonomously diagnose the issue and replan the subsequent path, demonstrating strong robustness in unseen scenarios.

 \item \textbf{A persistent memory mechanism drives continuous, experience-driven learning.} We introduce a long-term memory system that enables the agent to learn from operational experience. This solidifies successful adjustments into new workflows and attributes failures to heuristic knowledge, addressing the critical learning gap in current agentic workflows.

 \item \textbf{A comprehensive benchmark enables rigorous evaluation of agentic workflows.} To facilitate standardized evaluation, we construct a new benchmark comprising 162 distinct RS task types with 324 corresponding real-world workflows. This benchmark is specifically designed to rigorously assess the capability of an agent in planning workflow under given tool constraints.
\end{itemize}

The remainder of this paper is structured as follows. 
Section 2 provides a comprehensive review of the literature, covering general LLM-based agents, agentic workflows, and AI agents for Earth observation tasks. 
Section 3 introduces the proposed Cangling-KnowFlow framework and elaborates on its three key components. 
Section 4 outlines the constructed toolset database, the curated KnowFlow-Bench dataset, and the evaluation protocols. 
Section 5 reports the experimental settings and results, followed by an in-depth analysis. 
Finally, Section 6 concludes this paper and discusses future research directions.

\section{Related Work}\label{relatedwork}

\subsection{General-Purpose LLM Agents}\label{subsec:general_llm_agents}

The conceptual foundation of our research lies in the rapid evolution of general-purpose agents powered by LLMs. 
Early efforts demonstrated that LLMs could perform preliminary planning via techniques such as Chain-of-Thought prompting. 
A pivotal development was the introduction of frameworks like ReAct \cite{yao2023react}, which synergized reasoning and acting by interleaving thought processes with tool-based actions, enabling agents to interact with external environments to solve tasks. 
It further led to a wave of research into enhancing agent capabilities, particularly in generalized tool use, supported by the creation of comprehensive benchmarks and datasets \cite{li2023api, tang2023toolalpaca}. 
Building upon this, subsequent research has focused on enhancing agent autonomy and intelligence. 
For example, some approaches have incorporated self-reflection and memory mechanisms to allow agents to learn from past failures \cite{shinn2023reflexion}, while others have integrated tree-search algorithms to improve planning and decision-making capabilities \cite{zhou2024language}. 
Concurrently, multi-agent conversational frameworks have been proposed to tackle complex problems through collaborative task decomposition \cite{wu2024autogen}. 
Despite these advances, a common limitation of general-purpose agents is their reliance on the intrinsic, generalized knowledge of the underlying LLM. 
In specialized domains, this often leads to 'planning hallucinations'.
As highlighted by recent benchmarks \cite{liu2023agentbench}, these agents tend to generate plans that appear plausible but are scientifically incorrect.

\subsection{Agentic Workflows}\label{subsec:agentic_workflows}

To address the limitations of unstructured, general-purpose agents, the paradigm of the Agentic Workflow has emerged. 
This approach imposes a structured, often procedural, constraint on agent behavior to increase reliability and ensure compliance with domain-specific constraints. 
A key focus within this paradigm is the infusion of external knowledge to guide the planning process. 
Some studies have proposed augmenting agents with explicit knowledge bases to constrain the generation of action plans, thereby improving their factuality and relevance \cite{zhu2025knowagent, ou2025automind}. 
Others have explored the inherent tension between procedural compliance and operational flexibility, developing frameworks that allow agents to adhere to a predefined workflow while also managing unexpected, out-of-scope situations \cite{shi2025flowagent}. 
This relates directly to the need for dynamic adjustment in real-world applications. 
More recently, the concept of memory has been integrated into workflows, enabling agents to learn and reuse successful procedural patterns from past experiences \cite{wang2024agent}. 
Some advanced systems even attempt to automate the generation of the workflow itself \cite{zhang2024aflow}. 
However, the systematic definition, orchestration, dynamic adaptation, and experiential optimization of such workflows for highly specialized domains like RS remain open questions.

\subsection{AI Agents in RS and Geoscience}\label{subsec:ai_agents_rs}

The application of AI agents to RS and Geoscience is a nascent but rapidly developing field. 
Early efforts have focused predominantly on tool augmentation, demonstrating that agents can interpret user commands to invoke specific geospatial analysis tools or models. 
These include single-agent systems for tool execution \cite{xu2024rs, wei2025geotool, zhang2024geogpt, akinboyewa2025gis, chen2024llm, ning2024llm} and multi-agent frameworks for specialized data processing or collaborative analysis \cite{lin2025shapefilegpt, lee2025multi, tang2025geosr, pantiukhin2025accelerating, xie2025rag, yu2024mineagent}. 
While foundational, these systems are often limited to simple, single-turn tool calls and lack the capability to orchestrate complex, multi-step analytical workflows.

More recently, GeoFlow \cite{bhattaram2025geoflow} introduced the Agentic Workflow paradigm to the geospatial domain, representing a significant step forward. 
However, this and similar early attempts exhibit three critical deficiencies that are worth noting. 
First, they rely on the LLM's general knowledge for planning and lack a mechanism for injecting codified, expert-validated procedural knowledge, making them prone to domain-specific errors. 
Second, their workflows are largely static, planned once at the beginning of a task, and lack the robustness to dynamically adjust to runtime failures. 
Third, they are memoryless, precluding the ability to learn from or evolve through accumulated task logs and retrieved historical cases. 
Furthermore, while benchmarks for evaluating tool-use or specific spatial cognition abilities are emerging \cite{shabbir2025thinkgeo}, a comprehensive benchmark designed to rigorously assess an agent's advanced capabilities in complex workflow orchestration, dynamic adaptation, and empirical learning is still absent, which constitutes the motivation of this paper. 

\section{Methodology}\label{methodology}
\subsection{Framework Overview}\label{subsec:framework_overview}
\begin{figure*}[!t]
\centering
\includegraphics[width=1\linewidth]{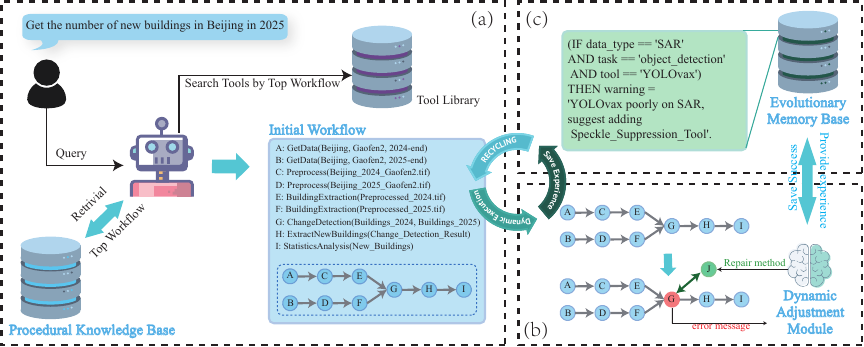}
\caption{The overall architecture of the CangLing-KnowFlow framework, illustrating the interactions between the Orchestrator Agent, the Procedural Knowledge Base, the Dynamic Execution Engine, and the Evolutionary Memory Module.}
\label{fig:framework}
\end{figure*}

The CangLing-KnowFlow framework is developed to fill the above-mentioned critical gaps by addressing the core challenge of fusing deep domain knowledge with a flexible, adaptive procedural flow for automating complex RS tasks.
The key design principle is to create an agent that not only executes predefined plans, but which also adapts dynamically to unforeseen circumstances and learns systematically from its operational experience. 
As shown in Figure \ref{fig:framework}, the architecture of our proposed CangLing-KnowFlow is composed of three primary, synergistic components, integrated under a central Orchestrator Agent. 
This agent serves as the cognitive core, responsible for interpreting user requests, planning the execution, and initiating interventions. 
It draws plans from an expert-informed Procedural Knowledge Base (PKB), a repository of codified, expert-validated RS workflows. 
The concrete execution is handled by a Dynamic Execution Engine, which interfaces with the toolset and provides real-time feedback on the outcome of each action. 
Finally, an Evolutionary Memory Module records all execution traces, enabling the agent to learn from experience by analyzing successes and failures to refine its knowledge over time.

The decision-making process of the Orchestrator Agent can be formally modeled as a Markov Decision Process (MDP) \cite{shi2025flowagent}. 
At each time step $t$, the agent selects an action $a_t$ based on the history of the current task and guidance from its knowledge sources.
We extend the traditional workflow agent formulation to incorporate the influence of long-term, cross-task experience explicitly. 
The action $a_t$ is determined by the agent's policy $\mathcal{A}$ as follows:

\begin{equation}
a_t \leftarrow \mathcal{A}(\mathcal{H}_{t-1}, W_{current}, \mathcal{K}_{evo}),
\label{eq:mdp}
\end{equation}
where $\mathcal{H}_{t-1}$ represents the execution history of the current task up to step $t-1$, including all previous actions and their observed outcomes. 
$W_{current}$ is the current workflow DAG, serving as the primary procedural guide. 
Critically, $\mathcal{K}_{evo}$ represents the generalized, evolutionary knowledge retrieved from the memory module, which comprises verified sub-plans and error-correction heuristics distilled from past task executions (details are provided in Section \ref{subsec:evolutionary_memory}). 
This term formally distinguishes our framework, as it allows the agent's decisions to be informed not just by its immediate context, but by its entire accumulated experience.

The practical implementation of the above formal model is realized through a structured operational cycle: (1) Task Decomposition and Workflow Retrieval, where the Orchestrator parses the goal of users and retrieves the most relevant workflow template from the PKB (see Section \ref{subsec:procedural_knowledge_base}). 
(2) Situated Planning, where the Orchestrator adapts the retrieved template to the specific data and contextual parameters of the current task. 
(3) Monitored Execution, where the Dynamic Execution Engine executes the planned steps sequentially, with the Orchestrator supervising the process. 
(4) Dynamic Adjustment, which is triggered upon execution failure, prompting the Orchestrator to diagnose the issue and re-plan the workflow. 
(5) Empirical Learning, which occurs post-task, where the Evolutionary Memory Module analyzes the complete execution trace to update the system's knowledge base. 
This operational cycle is formally detailed in Algorithm \ref{alg:agent_dynamic_learning}. 
The subsequent sections will provide a detailed discussion of the three components, i.e., the PKB, the dynamic adjustment mechanism, and the evolutionary memory module.

\begin{algorithm*}[!t]
 \caption{Dynamic Task Execution and Learning Algorithm}
 \label{alg:agent_dynamic_learning}
 \begin{algorithmic}[1] 
 \Require 
 \Statex User Task Goal $G_{user}$
 \Statex Procedural Knowledge Base $\mathcal{K}_{proc}$ (containing workflow templates)
 \Statex Evolutionary Memory Module $\mathcal{K}_{evo}$ (containing heuristic rules)
 \Statex Toolset $\mathcal{T}$
 \Ensure 
 \Statex Final Result $R_{final}$
 \Statex Updated Procedural Knowledge Base $\mathcal{K}'_{proc}$
 \Statex Updated Evolutionary Memory Module $\mathcal{K}'_{evo}$

 \Procedure{KnowFlowExecution}{$G_{user}, \mathcal{K}_{proc}, \mathcal{K}_{evo}, \mathcal{T}$}
 \Statex {\textbf{---Phase 1: Initialization and Planning---}}
 \State $W_{template} \gets \text{RetrieveWorkflowTemplate}(G_{user}, \mathcal{K}_{proc})$
 \State $W_{current} \gets \text{InstantiateWorkflow}(W_{template}, G_{user})$ \Comment{Adapt template to current context}
 \State $\mathcal{H} \gets \emptyset$ \Comment{Initialize execution history}
 \State $isSuccess \gets \text{false}$
 
 \Statex {\textbf{---Phase 2: Monitored Execution and Dynamic Adjustment---}}
 \While{not all nodes in $W_{current}$ are executed}
     \State $v_i \gets \text{GetNextExecutableNode}(W_{current})$
     \State $t_i \gets \text{GetToolForNode}(v_i, \mathcal{T})$
     \State $result_i, status_i \gets \text{ExecuteTool}(t_i, \text{inputs from } \mathcal{H})$
     \State $\mathcal{H} \gets \mathcal{H} \cup \{v_i, t_i, result_i, status_i\}$ \Comment{Update history}
     
     \If{$status_i$ is \textbf{not} "Success"}
         \Statex \hskip\algorithmicindent {\textit{--- Hierarchical Repair Strategy ---}}
         \State $repair\_action \gets \text{QueryMemoryForRepair}(\mathcal{H}, \mathcal{K}_{evo})$ \Comment{Tier 1: Knowledge-driven}
         \If{$repair\_action$ is $\emptyset$}
             \State $repair\_action \gets \text{LLMGenerateRepair}(\mathcal{H}, W_{current}, \mathcal{T})$ \Comment{Tier 2: LLM-driven}
         \EndIf
         
         \If{$repair\_action$ is \textbf{not} $\emptyset$}
             \State $W_{current} \gets \text{ApplyRepair}(W_{current}, v_i, repair\_action)$ \Comment{Manipulate DAG}
             \State \textbf{continue} \Comment{Continue execution from the repaired point}
         \Else
             \State $isSuccess \gets \text{false}$
             \State \textbf{break} \Comment{Terminal failure}
         \EndIf
     \EndIf
 \EndWhile
 
 \If{loop completed without break} 
    \State $isSuccess \gets \text{true}$ 
 \EndIf

 \Statex {\textbf{---Phase 3: Empirical Learning---}}
 \If{$isSuccess$ is $true$}
     \If{workflow was adjusted}
         \State $\mathcal{K}'_{proc} \gets \text{SolidifySuccess}(\mathcal{K}_{proc}, W_{current}, \mathcal{H})$ \Comment{Success Solidification}
     \EndIf
 \Else
     \State $\mathcal{K}'_{evo} \gets \text{AttributeFailure}(\mathcal{K}_{evo}, \mathcal{H})$ \Comment{Failure Attribution}
 \EndIf
 
 \State $R_{final} \gets \text{PackageResult}(\mathcal{H})$
 \State \Return $R_{final}, \mathcal{K}'_{proc}, \mathcal{K}'_{evo}$
 \EndProcedure
 \end{algorithmic}
\end{algorithm*}

\subsection{Expert-Informed Procedural Knowledge Base}\label{subsec:procedural_knowledge_base}
\begin{figure*}[!t]
\centering
\includegraphics[width=1\linewidth]{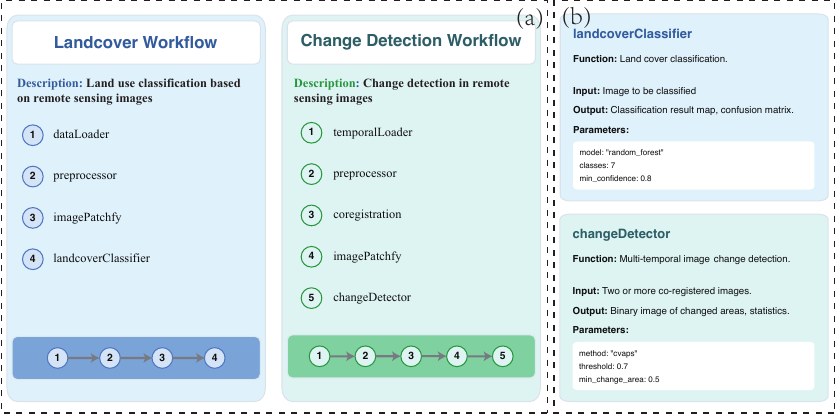}
\caption{Structure of the Procedural Knowledge Base (PKB). (a) An illustration of a hierarchical workflow template, showing the decomposition of a composite task into methods and atomic tool actions. (b) An example of the formalized tool schema for a single tool, detailing its parameters, preconditions, and effects.}
\label{fig:Tool_and_Workflow}
\end{figure*}

The PKB is the foundational component for grounding the agent's planning capabilities upon established scientific methodologies. 
It addresses the problem of planning hallucination by providing the Orchestrator Agent with a structured source of domain-specific procedural knowledge. 
The PKB is comprised of two core elements: a formalized Tool Schema and a library of hierarchical Workflow Templates, underpinned by a formal mathematical representation. 
The concrete structure of these elements is illustrated in Figure \ref{fig:Tool_and_Workflow}.

\subsubsection{Mathematical Formulation of Workflows}\label{subsubsec:mathematical_formulation}

To represent the structure and logic of RS analysis procedures formally, we model each workflow as a Directed Acyclic Graph (DAG). 
A workflow $W$ is defined as a tuple $W = (V, E)$, where $V$ is a set of vertices representing tool actions and $E$ is a set of directed edges representing dependencies.
Let $\mathcal{T} = \{t_1, t_2, \dots, t_n\}$ be the set of all available tools. 
Each vertex $v_i \in V$ corresponds to the invocation of a specific tool $t_i \in \mathcal{T}$. 
The properties of each tool $t_i$ are mainly defined by its two predefined mapping functions, precondition $\text{Pre}(t_i)$, and its execution effects, $\text{Eff}(t_i)$. 
The set of edges is constructed based on semantic compatibility, with
a directed edge $e_{ij} = (v_i, v_j) \in E$ exists from vertex $v_i$ to $v_j$ if and only if an effect of tool $t_i$ satisfies a precondition of tool $t_j$, defined as
\begin{equation}
    E=\{e_{ij}|\phi_{LLM}(\text{Eff}(t_i), \text{Pre}(t_j))=1\},
\end{equation}
where $\phi_{LLM}$ serves as a boolean indicator function implemented by an LLM to determine if $\text{Eff}(t_i)$ conceptually entails $\text{Pre}(t_j)$.

The acyclic nature of the graph ensures that a workflow is logically sound and executable. 
This formal DAG representation plays a central role in the CangLing-KnowFlow framework, as it enables the agent to check the logical consistency of the workflow, identify failure points, and systematically perform dynamic adjustments by manipulating the graph structure.
In contrast with existing LLM-based agents, the proposed checking manner follows certain practical rules, such as topological sorting, cycle detection, and computing logics, rather than relying purely on probabilities.

\subsubsection{PDDL Tool Schema}\label{subsubsec:pddl_tool_schema}

To enable the mathematical formulation, every tool available to the agent is described formally using a schema inspired by the Planning Domain Definition Language (PDDL), as shown in Figure \ref{fig:Tool_and_Workflow}b. Each tool $t \in \mathcal{T}$ is represented as an action with a clearly defined set of the following three components: 
(1) Parameters, define the input arguments for the tool, including data types and constraints;
(2) Preconditions, denoted as $\text{Pre}(t)$, record logical conditions that must be met before tool execution;
(3) Effects, denoted as $\text{Eff}(t)$), evaluate the state changes that occur upon tool execution.

We formalize the operational space of the agent, denoted as $\mathbf{\Omega}$, as a structured hierarchical taxonomy rather than a simple collection of functions, which aligns with the intrinsic multi-level nature of RS tasks. 
Specifically, $\mathbf{\Omega}$ is organized into two categories: 
(1) Atomic operators, which handle low-level data manipulation (e.g., geometric correction, projection); 
(2) Semantic operators, which encapsulate complex analysis tasks (e.g., segmentation, inversion) into high-level function calls.
This rigorous taxonomy ensures that the derived tool space $\mathcal{T} \subset \mathbf{\Omega}$ provides the necessary granularity for domain-specific benchmarks.
By unifying these heterogeneous functions into a machine-readable library, we establish the fundamental building blocks for constructing our workflow DAGs.
We will further detail the concrete instantiation of this toolset, including specific GDAL and deep learning algorithms, in Section \ref{toolset_formalization}.

\subsubsection{Workflow Templates}\label{subsubsec:hierarchical_templates}

The second element of the PKB is a library of workflow templates, which codify mature and complex RS analysis procedures as predefined DAGs (Figure \ref{fig:Tool_and_Workflow}a). 
These templates are structured hierarchically to represent tasks at multiple levels of abstraction. 
For example, a high-level (defined as conceptually abstract and goal-oriented) composite task, such as Regional Flood Inundation Assessment, is decomposed into a sequence of lower-level processing steps, such as Data Acquisition and Preprocessing. 
Each step is itself a sub-graph composed of concrete tool actions.

We designed this hierarchical structure to enable efficient, top-down planning.
By allowing the agent to retrieve and instantiate high-level templates for common tasks, it reduces the planning search space and ensures adherence to expert-validated procedures. 
This structural guidance constrains the generative process of the LLM, directing it to produce plans that are both syntactically correct and compliant with standard RS protocols.

\begin{figure*}[!t]
\centering
\includegraphics[width=1\linewidth]{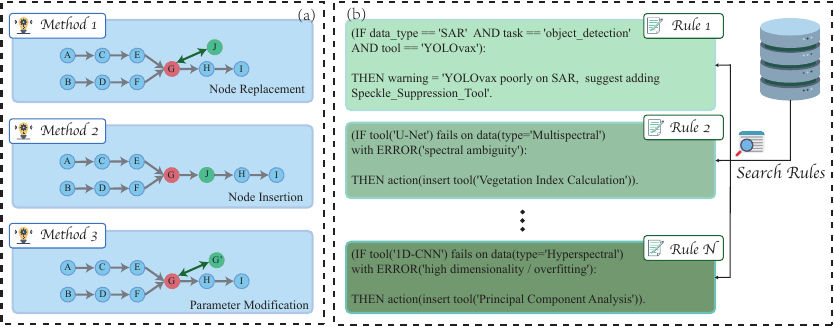}
\caption{Schematic of the Dynamic Workflow Adjustment and Evolutionary Memory. (a) The three graph manipulation operators used for workflow repair: Node Replacement, Node Insertion, and Parameter Modification. (b) The Evolutionary Memory, which stores heuristic Pattern-Action rules distilled from past experiences to guide the adjustment process.}
\label{fig:adjustment}
\end{figure*}

\subsection{Dynamic Workflow Adjustment}\label{subsec:dynamic_adjustment}
One key innovation of the CangLing-KnowFlow framework is its ability to adjust the workflow in response to runtime failures dynamically, thereby overcoming the inflexibility of static plans. 
This mechanism is the practical realization of the agent's policy $\mathcal{A}$ from Eq. (\ref{eq:mdp}) under conditions of uncertainty. 
In RS workflows, the source of this uncertainty could be complicated, such as ambiguities in the user's prompts (e.g., vague spatial definitions), intrinsic data irregularities (e.g., unexpected cloud cover or sensor noise), and unexpected tool behaviors (e.g., non-convergence of neural networks). 
When a failure occurs for the abovementioned reasons, the agent must compute a new action $a_t$ that deviates from the original plan. 
This is achieved through a structured process composed of three stages: failure detection, a hierarchical repair strategy, and workflow graph manipulations.

\subsubsection{Failure Detection and Diagnosis}\label{subsubsec:failure_detection}

The dynamic adjustment process is initiated when the Dynamic Execution Engine detects a failure at a specific vertex $v_f \in V$ of the current workflow $W_{current}$.
We define a failure as either an explicit error returned by a tool or an output that fails to meet a predefined quality metric. 
Once detected, the engine immediately halts the execution of dependent downstream nodes and reports the failure state, which becomes part of the history $\mathcal{H}_{t-1}$, to the Orchestrator Agent.
The Orchestrator then diagnoses the root cause by analyzing the history in conjunction with the preconditions of the failed tool $t_f$, aiming to identify the root cause of the failure.

\subsubsection{Hierarchical Repair Strategy}\label{subsubsec:hierarchical_repair}

Once a failure is diagnosed, the Orchestrator employs a hierarchical strategy to generate a repair plan.
Specifically, it selects the next action $a_t$ by prioritizing the knowledge from past experience $\mathcal{K}_{evo}$ over generalized reasoning.
\begin{itemize}
 \item \textbf{Tier 1: Knowledge-driven Repair.} The agent first queries the Evolutionary Memory Module for $\mathcal{K}_{evo}$ related to similar failures. If a previously successful repair strategy is found, it is immediately adopted as the next action $a_t$. This approach leverages past experience to solve recurring problems quickly and reliably.
 \item \textbf{Tier 2: LLM-driven Creative Repair.} If no relevant solution is found in $\mathcal{K}_{evo}$, the agent pushes the problem to the LLM. It provides the LLM with the full context ($\mathcal{H}_{t-1}$, $W_{current}$) to generate a novel repair plan, which then defines the action $a_t$.
\end{itemize}

\subsubsection{Workflow Graph Manipulation}\label{subsubsec:graph_manipulation}

The repair plan, formalized as the chosen action $a_t$, is executed by applying a series of topological transformations to the current workflow DAG $W_{current}$, resulting in an updated workflow structure $W'_{current}$. 
These transformations primarily consist of three distinct graph operations designed to rectify the identified fault. 
The first transformation, Node Replacement, involves substituting the failed vertex $v_f$ with a new vertex $v_r$ associated with an alternative tool $t_r$, ensuring that all original incoming and outgoing edges of $v_f$ are correctly remapped to $v_r$.
Alternatively, the agent may perform Node Insertion, which introduces a new intermediate vertex $v_{new}$ between a predecessor node $v_p$ and the failed node $v_f$. 
This operation structurally modifies the path by severing the direct link $(v_p, v_f)$ and establishing a new sequence of edges $(v_p, v_{new})$ and $(v_{new}, v_f)$. 
Finally, Parameter Modification allows for in-situ adjustment, where the vertex $v_f$ is topologically retained, but the hyperparameters governing its underlying tool $t_f$ are recalibrated to resolve the execution error.

\subsection{Experience-driven Memory and Evolution}\label{subsec:evolutionary_memory}
The Evolutionary Memory Module plays a role that endows the CangLing-KnowFlow agent with the ability to learn and evolve from its operational experience, addressing the critical learning gap in existing agentic workflows. 
It is mainly achieved through a post-task analysis and knowledge generalization process that operates on a repository of detailed execution traces.

\subsubsection{Structure of the Evolutionary Memory}\label{subsubsec:memory_structure}

The memory module serves as the persistent storage for the evolutionary knowledge $\mathcal{K}_{evo}$ introduced in Eq. (\ref{eq:mdp}). 
Structurally, it is organized as a database of execution traces. 
Each trace, $\mathcal{E}$, is a comprehensive log of a completed task, containing the initial user request, the initial workflow DAG $W_{init}$, the full sequence of actions taken, all intermediate data and tool outputs, any failures encountered, the dynamic adjustments performed, and the final outcome. 
This rich and structured dataset serves as the solid foundation for the learning processes of our agent.

\subsubsection{Knowledge Generalization and Refinement}\label{subsubsec:knowledge_generalization_and_refinement}

After a task is completed, the Orchestrator Agent initiates a knowledge generalization phase, which involves two distinct learning pathways depending on the task outcome:

Success Solidification (Positive Feedback): If a task that involved dynamic adjustments is completed successfully, the final validated workflow $W_{final}$ can then be considered a candidate worth promoting. 
After the agent analyzes the trace to identify the specific sequence of adjustments that led to success, 
the successful, dynamically generated workflow is then generalized into a new, reusable template. 
This template, along with metadata describing the context in which it was effective, will be finally added to the PKB. 
In this manner, the whole process could continuously enrich the PKB with novel, field-tested procedures.

Failure Attribution (Negative Feedback): In the event of a terminal task failure, the agent performs a failure attribution analysis on the execution trace $\mathcal{E}$. 
It identifies the specific failure point $v_f$, the state of the workflow at that time, and the diagnosed root cause. 
Based on this analysis, it synthesizes a new \textit{Pattern-Action Rule}, formally denoted as $r \in \mathcal{K}_{evo}$. 
This rule represents a conditional logic that maps a specific failure pattern to a recommended repair action:

\begin{quote}
\texttt{IF tool('YOLOvax') fails on data(type='SAR') with error('low\_contrast')} \\
\texttt{THEN Apply action(insert\_tool('Speckle\_\\Suppression'))}
\end{quote}

These generated rules are stored in a dedicated section of the memory module and are prioritized during the ``Knowledge-driven Repair" tier of the dynamic adjustment process. 
This creates a negative feedback loop that allows the agent to learn from its mistakes and provide a robust repair strategy when errors occur, significantly reducing the latency and token costs associated with repetitive LLM querying.

Through the dual mechanism of solidifying successes into templates and distilling failures into repair rules, the CangLing-KnowFlow agent can systematically transform its episodic experiences into durable procedural knowledge ($\mathcal{K}_{evo}$), enabling it to evolve and increase its accuracy over time.

\section{Database and Benchmark}\label{datasets}

To ground our agent's capabilities and rigorously evaluate its performance, we constructed two key databases: (1) an internal PKB to provide the agent with domain-specific knowledge and procedural guidance, and (2) a comprehensive evaluation suite, termed KnowFlow-Bench, to assess its workflow orchestration capabilities against expert-defined standards.

\begin{figure}[!t]
\centering
\includegraphics[width=0.8\linewidth]{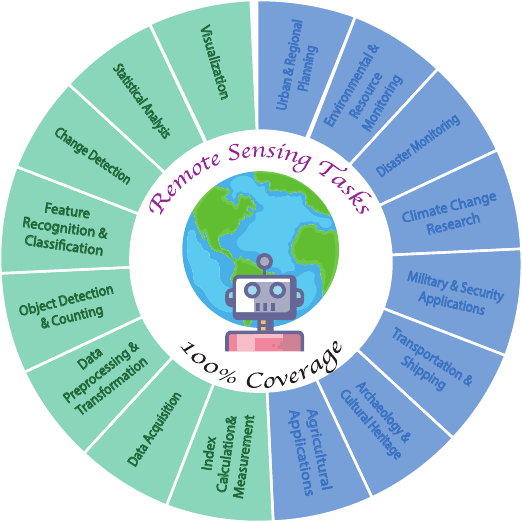}
\caption{CangLing-KnowFlow provides comprehensive support for 162 diverse remote sensing tasks, ranging from simple visualization to complex urban and regional planning.}
\label{fig:task_support}
\end{figure}

\subsection{Procedural Knowledge Base Construction}

The PKB is the cornerstone of the CangLing-KnowFlow framework, designed to infuse the agent with expert-level procedural knowledge. 
This section details the practical construction of the PKB, whose formal structure and mathematical preliminaries were defined in Section \ref{subsec:procedural_knowledge_base}. 
The construction involved a two-stage process of toolset formalization and high-level workflow curation.

\subsubsection{Toolset Formalization}\label{toolset_formalization}

The agent's operational capabilities are defined by a comprehensive toolset. 
Each tool was formalized according to the PDDL-like schema specified in Section \ref{subsubsec:pddl_tool_schema}. 
The curated toolset is categorized into traditional and deep learning-based tools to cover a wide spectrum of RS applications.

Traditional Tools: We incorporated the complete suite of traditional data processing algorithms from the Geospatial Data Abstraction Library (GDAL). 
Each algorithm was formalized as a distinct tool, providing exhaustive coverage of fundamental geospatial data manipulation operations.

Deep Learning Tools: We designed a suite of high-level deep learning tools, each tailored to a primary RS analysis task.
These tools decouple functional logic from architectural complexity, allowing the system to focus on the transformation of input data into analytical outputs.
The suite includes four main categories: semantic\_segmentation, object\_detection, change\_detection, and remote\_sensing\_inversion. 
Each tool is defined by a rigorous JSON schema that specifies its functionality, parameters, and input/output requirements, ensuring reliable interpretation and invocation by the agent.
For example, the semantic\_segmentation tool requires parameters such as image\_path, target\_classes, and spatial\_resolution\_m to execute pixel-wise classification.

\subsubsection{High-Level Workflow Curation}

To populate the library of hierarchical workflow templates described in Section \ref{subsubsec:hierarchical_templates}, we curated a large-scale collection of high-level procedural guides. 
The process began by generating a diverse set of 162 realistic RS task descriptions using a state-of-the-art LLM (Gemini 2.5 Pro) \cite{comanici2025gemini}. 
For each task, domain experts then manually annotated between two and 10 plausible, high-level solution workflows, resulting in a total of 1,008 unique templates. 
These templates are not rigid, function-call-level scripts; rather, they are abstract procedural sequences (e.g., Step 1: perform radiometric correction, Step 2: classify land cover) that embody the hierarchical structure defined in our methodology. 
This library provides the agent with a rich source of expert-derived strategies, serving as the primary input for its situated planning phase.

\subsection{Evaluation Benchmarks}

A rigorous evaluation of agentic capabilities necessitates well-designed benchmarks. 
While general-purpose benchmarks have been instrumental in advancing the field \cite{li2023api, liu2023agentbench}, they often lack the domain specificity required for scientific applications. 
Conversely, existing benchmarks in the geospatial domain tend to focus on discrete capabilities rather than the holistic process of workflow orchestration. 
To address this gap, we developed a new, comprehensive benchmark, KnowFlow-Bench, and additionally evaluated our agent on the existing ThinkGeo benchmark \cite{shabbir2025thinkgeo} to test its generalizability.

\subsubsection{The KnowFlow-Bench}

Existing benchmarks typically focus on isolated capabilities,
for example, ThinkGeo \cite{shabbir2025thinkgeo} assesses step-by-step tool usage, while GeoCode-Bench \cite{hou2025can} centers specifically on geospatial code generation. 
Consequently, the field lacks a dedicated standard for evaluating an agent's comprehensive capacity to translate high-level goals into executable workflows.

To bridge this gap, KnowFlow-Bench is built upon the 162 tasks curated from our PKB. 
For each task, domain experts annotated two distinct, complete, and executable ground-truth workflows, yielding a total of 324 evaluation instances. 
Unlike the high-level templates in the PKB, each workflow in KnowFlow-Bench consists of a concrete sequence of function calls with fully specified parameters.
These ``gold standard" workflows represent scientifically valid solution paths given the available toolset, serving as the reference for evaluating both the procedural correctness and final output accuracy of the agent.

\subsubsection{Cross-Benchmark Generalization on ThinkGeo}

We also extended our evaluation to the ThinkGeo benchmark \cite{shabbir2025thinkgeo} to test the transferability and generalization of our framework.
ThinkGeo is a comprehensive agentic benchmark comprising 436 tasks grounded in real-world RS imagery, encompassing seven major domains such as urban planning and disaster assessment. 
It provides a suite of 14 predefined tools and evaluates agents based on their ability to follow a ReAct-style reasoning and execution chain. 
To facilitate this cross-evaluation, we configured the CangLing-KnowFlow agent to align with the ThinkGeo environment.
This adaptation entailed extracting the procedural workflow logic from its ReAct-style annotations and restricting the agent's action space strictly to the 14 tools provided by the benchmark. 
This set of experiments highlights the robustness of our agent's planning mechanisms, confirming its effectiveness even when operating with external tool definitions and novel task requirements.

\subsection{Evaluation Metrics}
\label{subsec:evaluation_metrics}

To ensure a holistic assessment of agent performance, we designed a comprehensive evaluation protocol that rigorously evaluates both the final task outcome and the procedural fidelity, logical coherence, and step-wise accuracy of the execution trace.
Accordingly, our protocol is organized into two primary categories, end-to-end outcome evaluation and step-wise process evaluation.

\subsubsection{End-to-End Evaluation}

The first category assesses the overall effectiveness and efficiency of an agent in accomplishing a given RS task from start to finish. 
It is used to compare the performance of different agentic frameworks (e.g., ReAct, Reflexion, CangLing-KnowFlow) across our benchmarks.

\textbf{Task Success Rate (TSR)}: This is the primary metric for overall performance. 
A task is considered successful if the agent completes the workflow and produces a final result that is semantically and numerically correct when compared to the ground-truth, with minimal human intervention.

\textbf{First-Pass Accuracy (FPA)}: The proportion of tasks that are successfully completed using the initially generated workflow, without requiring any dynamic adjustments. 
This metric specifically measures the quality of the agent's initial planning.

\textbf{Planning Efficiency}: This set of metrics quantifies the cost and autonomy of the agent's planning process, primarily consists of (1)
\textbf{Number of Tool Calls (NTC)}: The total count of tool invocations required to complete the task. A smaller number generally indicates a more efficient plan, and (2)
\textbf{Number of Interactions (NI)}: The frequency of required user interventions or confirmations. A lower count signifies greater autonomy.

\subsubsection{Step-wise Execution Evaluation}

The second category provides a fine-grained analysis of the procedural correctness of our CangLing-KnowFlow agent's execution trace. 
To achieve this, we adopt the fine-grained, step-wise evaluation metrics proposed in the GTA benchmark \cite{wang2024gta}. 
This manner is designed to assess the procedural fidelity of an agent's reasoning and action trace at each step of the workflow, offering deep insights into its operational capabilities.

\textbf{Instruction Following Accuracy (InstAcc)}: This metric evaluates whether the agent's output at each step follows strictly to the predefined format (e.g., Thought: ... Action: ...). 
It serves as a fundamental check of the model's ability to generate structured output.

\textbf{Tool Selection Accuracy (ToolAcc)}: It measures whether the tool selected by the agent at a given step matches the one in the ground-truth workflow, which is a core metric for evaluating the agent's task decomposition and planning abilities.

\textbf{Argument Correctness Accuracy (ArgAcc)}: Conditional on the correct tool being selected, this metric assesses the correctness of the parameters supplied to the tool. 
As highlighted by \cite{wang2024gta}, accurate argument prediction is often a critical bottleneck for agent performance. 
It is quantified by evaluating the exact correspondence between the assigned parameter values (including referenced variable names) and the ground-truths.

\textbf{Overall Score}: Calculated as the arithmetic mean of InstAcc, ToolAcc, and ArgAcc, which offers a unified quantitative measure of step-wise procedural fidelity.

\section{Experiments and Discussion}\label{experiments}
\subsection{Experimental Setup}
To rigorously evaluate the performance and efficacy of our proposed CangLing-KnowFlow framework, we designed a comprehensive experimental protocol. 
This section details the evaluation benchmarks, the selected baseline methods for comparison, the suite of LLMs used as backbones, and the specific metrics employed for quantitative assessment.

As discussed before, the evaluation was conducted on two complementary benchmarks: our developed KnowFlow-Bench and the public ThinkGeo benchmark \cite{shabbir2025thinkgeo}. 
The KnowFlow-Bench is tailored to assess an agent's end-to-end capabilities in orchestrating complex, multi-step RS workflows, from high-level task planning and tool invocation to dynamic adjustments in response to execution failures.
The ThinkGeo is used to further test the generalization capabilities and adaptability of our framework, allowing us to assess CangLing-KnowFlow's performance in a novel environment with a distinct set of predefined tools and task structures.

Given the lack of competitors in this emerging field, we established the following settings as baselines for comparison:
\begin{itemize}
 \item \textbf{LLM + Tools}: A foundational zero-shot baseline that relies solely on the raw, in-context learning ability of an LLM to interpret a user request and invoke the appropriate tools without any structured reasoning framework.

 \item \textbf{ReAct} \cite{yao2023react}: A foundational paradigm that synergizes reasoning and acting by generating interleaved ``Thought'' and ``Action'' traces. 
 It serves as the standard baseline, employing standard prompting strategies and the shared toolset. 
 This setup establishes performance metrics in the absence of external procedural knowledge augmentation.

 \item \textbf{Reflexion} \cite{shinn2023reflexion}: An advanced evolution of ReAct that incorporates a verbal reinforcement learning mechanism.
 When a failure is detected, the agent can analyze its execution trace to generate textual self-reflections, which act as heuristic feedback for subsequent trials. 
 While equipped with the same tools as the other agents, this baseline enables us to assess the efficacy of purely linguistic self-correction compared to our structured, graph-based evolutionary memory.

 \item \textbf{Ours (CangLing-KnowFlow)}: Our proposed framework, which integrates an expert-informed knowledge base with dynamic workflow adjustment and long-term evolutionary learning.

\end{itemize}

To ensure the robustness of our analysis and mitigate potential biases inherent to specific architectures, we evaluated our framework across a diverse family of state-of-the-art LLMs. 
As detailed in Table \ref{tab:performance_comparison}, our selection encompasses both leading proprietary models (e.g., GPT-4 \cite{achiam2023gpt}, Claude 3.7 Sonnet) and high-performance open-weight models (e.g., Llama3-70B \cite{grattafiori2024llama}). 
All evaluations strictly follows the comprehensive protocol defined in Section \ref{subsec:evaluation_metrics} to guarantee a holistic assessment.

\subsection{Overall Performance Comparison}
To validate the effectivity of our framework empirically, we conducted a comprehensive comparison against established baseline methods on our proposed KnowFlow-Bench. 
The detailed results summarized in Table \ref{tab:performance_comparison} offer a clear and consistent tendency that the CangLing-KnowFlow framework wins a significant improvement across all evaluated metrics and LLM backbones.

\begin{table*}[!t]
\centering
\caption{Performance comparison of different agent frameworks on KnowFlow-Bench. For each metric, the first value represents performance on simple tasks, and the second value (in parentheses) represents performance on complex tasks. Models are listed with their release year and availability status. The best result in each metric is highlighted in \textbf{bold}. `↑`=higher is better, `↓`=lower is better.}
\label{tab:performance_comparison}
\begin{tabular}{lllccccc}
\toprule
\textbf{LLM Backbone} & \textbf{Year} & \textbf{Type} & \textbf{Metric} & \textbf{LLM + Tools} & \textbf{ReAct}\cite{yao2023react} & \textbf{Reflexion}\cite{shinn2023reflexion} & \textbf{Ours} \\
\midrule
\multirow{4}{*}{\textbf{Mistral-7B}~\cite{jiang2023mistral}} & \multirow{4}{*}{2023} & \multirow{4}{*}{Open}
& TSR (\%) ↑ & 54.6 (38.9) & 58.9 (43.6) & 60.2 (45.1) & \textbf{68.4 (54.2)} \\
& & & FPA (\%) ↑ & 31.8 (21.3) & 35.1 (24.8) & 36.7 (26.1) & \textbf{45.9 (34.7)} \\
& & & NTC ↓ & 31.2 (42.6) & 28.3 (38.7) & 27.4 (37.2) & \textbf{19.6 (26.9)} \\
& & & NI ↓ & 14.7 (19.4) & 12.8 (16.8) & 11.9 (15.9) & \textbf{6.2 (9.7)} \\
\cmidrule(l){4-8}
\multirow{4}{*}{\textbf{Llama3-8B}~\cite{grattafiori2024llama}} & \multirow{4}{*}{2024} & \multirow{4}{*}{Open} 
& TSR (\%) ↑ & 58.3 (42.1) & 62.1 (46.8) & 63.7 (48.2) & \textbf{71.5 (57.9)} \\
& & & FPA (\%) ↑ & 35.2 (24.6) & 38.4 (28.1) & 40.1 (29.7) & \textbf{49.3 (38.4)} \\
& & & NTC ↓ & 28.4 (38.7) & 25.7 (34.9) & 24.8 (33.2) & \textbf{17.2 (23.8)} \\
& & & NI ↓ & 12.3 (16.8) & 10.8 (14.2) & 9.9 (13.1) & \textbf{4.8 (7.6)} \\
\cmidrule(l){4-8}
\multirow{4}{*}{\textbf{Llama3-70B}~\cite{grattafiori2024llama}} & \multirow{4}{*}{2024} & \multirow{4}{*}{Open}
& TSR (\%) ↑ & 73.4 (60.4) & 76.2 (64.2) & 77.1 (65.1) & \textbf{83.6 (74.6)} \\
& & & FPA (\%) ↑ & 49.7 (35.7) & 52.3 (39.3) & 53.4 (40.4) & \textbf{63.1 (51.1)} \\
& & & NTC ↓ & 21.5 (29.5) & 19.6 (26.6) & 18.8 (25.8) & \textbf{11.7 (17.7)} \\
& & & NI ↓ & 8.2 (13.2) & 7.1 (11.1) & 6.6 (10.6) & \textbf{2.8 (5.8)} \\
\cmidrule(l){4-8}
\multirow{4}{*}{\textbf{DeepSeek V3.1}~\cite{liu2024deepseek}} & \multirow{4}{*}{2024} & \multirow{4}{*}{Open}
& TSR (\%) ↑ & 74.8 (62.8) & 78.1 (67.1) & 79.3 (68.7) & \textbf{86.7 (78.3)} \\
& & & FPA (\%) ↑ & 51.2 (38.2) & 54.9 (42.9) & 56.4 (44.4) & \textbf{66.1 (53.8)} \\
& & & NTC ↓ & 21.3 (29.2) & 18.7 (25.8) & 17.9 (24.9) & \textbf{11.2 (16.4)} \\
& & & NI ↓ & 7.8 (11.6) & 6.3 (9.5) & 5.7 (8.8) & \textbf{2.9 (4.9)} \\
\cmidrule(l){4-8}
\multirow{4}{*}{\textbf{DeepSeek-R1}~\cite{guo2025deepseek}} & \multirow{4}{*}{2025} & \multirow{4}{*}{Open}
& TSR (\%) ↑ & 83.4 (77.4) & 86.9 (82.9) & 87.8 (84.8) & \textbf{93.1 (89.1)} \\
& & & FPA (\%) ↑ & 62.7 (54.7) & 66.3 (59.3) & 67.9 (61.9) & \textbf{75.4 (69.4)} \\
& & & NTC ↓ & 16.2 (20.2) & 14.1 (17.1) & 13.4 (16.4) & \textbf{8.9 (10.9)} \\
& & & NI ↓ & 5.9 (8.9) & 4.7 (6.7) & 4.2 (6.2) & \textbf{1.8 (2.8)} \\
\cmidrule(l){4-8}
\multirow{4}{*}{\textbf{GPT-3.5-Turbo}} & \multirow{4}{*}{2022} & \multirow{4}{*}{Commercial}
& TSR (\%) ↑ & 68.1 (50.1) & 70.3 (54.3) & 71.8 (55.8) & \textbf{79.2 (65.2)} \\
& & & FPA (\%) ↑ & 42.3 (28.3) & 44.8 (31.8) & 46.1 (33.1) & \textbf{57.4 (43.4)} \\
& & & NTC ↓ & 23.7 (35.7) & 21.8 (32.8) & 20.9 (31.9) & \textbf{14.3 (21.3)} \\
& & & NI ↓ & 9.8 (16.8) & 8.6 (14.6) & 7.9 (13.9) & \textbf{3.5 (7.5)} \\
\cmidrule(l){4-8}
\multirow{4}{*}{\textbf{GPT-4}~\cite{achiam2023gpt}} & \multirow{4}{*}{2023} & \multirow{4}{*}{Commercial}
& TSR (\%) ↑ & 79.7 (69.7) & 82.9 (74.9) & 84.5 (76.5) & \textbf{89.1 (84.1)} \\
& & & FPA (\%) ↑ & 57.3 (45.3) & 61.2 (50.2) & 62.8 (51.8) & \textbf{69.8 (61.8)} \\
& & & NTC ↓ & 17.8 (24.8) & 15.6 (21.6) & 14.7 (20.7) & \textbf{9.3 (13.3)} \\
& & & NI ↓ & 6.9 (10.9) & 5.6 (8.6) & 5.1 (8.1) & \textbf{1.8 (2.8)} \\
\cmidrule(l){4-8}
\multirow{4}{*}{\textbf{Gemini 1.5 Pro}~\cite{team2024gemini}} & \multirow{4}{*}{2024} & \multirow{4}{*}{Commercial}
& TSR (\%) ↑ & 78.1 (66.1) & 81.3 (71.3) & 82.6 (72.6) & \textbf{87.9 (81.9)} \\
& & & FPA (\%) ↑ & 55.6 (42.6) & 59.1 (47.1) & 60.8 (48.8) & \textbf{67.3 (58.3)} \\
& & & NTC ↓ & 18.6 (26.6) & 16.7 (23.7) & 15.8 (22.8) & \textbf{10.1 (15.1)} \\
& & & NI ↓ & 7.3 (11.3) & 5.9 (8.9) & 5.4 (8.4) & \textbf{2.1 (3.1)} \\
\cmidrule(l){4-8}
\multirow{4}{*}{\textbf{Claude 3.5 Sonnet}} & \multirow{4}{*}{2024} & \multirow{4}{*}{Commercial}
& TSR (\%) ↑ & 82.1 (75.1) & 85.4 (80.4) & 86.8 (81.8) & \textbf{90.2 (87.2)} \\
& & & FPA (\%) ↑ & 58.3 (49.3) & 62.1 (54.1) & 63.7 (55.7) & \textbf{71.5 (65.5)} \\
& & & NTC ↓ & 16.2 (22.2) & 14.5 (19.5) & 13.9 (18.9) & \textbf{9.1 (12.1)} \\
& & & NI ↓ & 6.1 (9.1) & 4.8 (6.8) & 4.3 (6.3) & \textbf{1.7 (2.7)} \\
\cmidrule(l){4-8}
\multirow{4}{*}{\textbf{Claude 3.7 Sonnet}} & \multirow{4}{*}{2024} & \multirow{4}{*}{Commercial}
& TSR (\%) ↑ & 84.3 (78.3) & 87.6 (83.6) & 88.9 (84.9) & \textbf{92.4 (90.4)} \\
& & & FPA (\%) ↑ & 61.2 (53.2) & 65.0 (58.0) & 66.4 (59.4) & \textbf{74.1 (69.1)} \\
& & & NTC ↓ & 15.3 (20.3) & 13.2 (17.2) & 12.6 (16.6) & \textbf{8.3 (10.3)} \\
& & & NI ↓ & 5.5 (8.5) & 4.2 (6.2) & 3.8 (5.8) & \textbf{1.3 (2.3)} \\
\cmidrule(l){4-8}
\multirow{4}{*}{\textbf{GPT-5}} & \multirow{4}{*}{2025} & \multirow{4}{*}{Commercial}
& TSR (\%) ↑ & 86.8 (82.8) & 89.9 (87.9) & 91.1 (89.1) & \textbf{94.7 (93.7)} \\
& & & FPA (\%) ↑ & 64.5 (58.5) & 68.1 (63.1) & 69.6 (64.6) & \textbf{76.8 (73.8)} \\
& & & NTC ↓ & 14.1 (18.1) & 12.3 (15.3) & 11.7 (14.7) & \textbf{7.6 (8.6)} \\
& & & NI ↓ & 4.9 (7.9) & 3.7 (5.7) & 3.3 (5.3) & \textbf{1.0 (1.0)} \\
\cmidrule(l){4-8}
\multirow{4}{*}{\textbf{Gemini-2.5-Pro}~\cite{comanici2025gemini}} & \multirow{4}{*}{2025} & \multirow{4}{*}{Commercial}
& TSR (\%) ↑ & 85.7 (81.7) & 88.9 (86.9) & 90.1 (88.1) & \textbf{95.3 (92.3)} \\
& & & FPA (\%) ↑ & 64.2 (56.2) & 67.8 (61.8) & 69.4 (63.4) & \textbf{76.9 (71.9)} \\
& & & NTC ↓ & 15.4 (19.4) & 13.2 (16.2) & 12.6 (15.6) & \textbf{7.8 (9.8)} \\
& & & NI ↓ & 5.1 (7.3) & 3.9 (5.9) & 3.5 (5.5) & \textbf{1.2 (2.2)} \\
\cmidrule(l){4-8}
\multirow{4}{*}{\textbf{Claude 4 Sonnet}} & \multirow{4}{*}{2025} & \multirow{4}{*}{Commercial}
& TSR (\%) ↑ & 88.7 (84.7) & 91.5 (89.5) & 92.6 (90.6) & \textbf{96.1 (95.1)} \\
& & & FPA (\%) ↑ & 67.1 (61.1) & 70.4 (66.4) & 71.7 (67.7) & \textbf{79.2 (76.2)} \\
& & & NTC ↓ & 13.2 (17.2) & 11.1 (14.1) & 10.5 (13.5) & \textbf{6.8 (7.8)} \\
& & & NI ↓ & 4.3 (7.3) & 3.1 (5.1) & 2.7 (4.7) & \textbf{0.8 (0.8)} \\
\bottomrule
\end{tabular}%
\end{table*}

Most notably, the TSR metric highlights the distinct advantage of our proposed approach.
Across both simple and complex tasks, CangLing-KnowFlow consistently achieves the highest success rates, outperforming baselines by significant margins.
For example, when powered by the Claude 4 Sonnet model, our framework boosts the TSR on complex tasks to 95.1\%, a substantial improvement over the 90.6\% achieved by the nearest competitor, Reflexion. 
This result validates the efficacy of integrating an expert-informed PKB, which guides the agent with scientifically valid prior knowledge, thereby effectively mitigating the logical hallucinations and inefficient planning often observed in unconstrained agent architectures.

Furthermore, the results highlight the framework's capability to significantly enhance planning efficiency and autonomy.
CangLing-KnowFlow consistently yields higher FPA scores than all baselines, indicating that initial workflows derived from our knowledge base are inherently robust and less susceptible to early-stage failure.
Consequently, this leads to a substantial reduction in both the NTC and the NI. 
By offering a validated, direct path to the solution, our approach curtails the redundant iterative loops that are typical in ReAct-based agents, thereby lowering computational costs and enhancing the degree of automation.

\begin{table*}[!t]
\centering
\caption{Generalization performance on the ThinkGeo benchmark, evaluated using the four primary metrics from our main experiments. The results consistently show that CangLing-KnowFlow's architectural advantages in success rate, planning quality, and efficiency are transferable to novel environments. Best results are in \textbf{bold}. `↑`=higher is better, `↓`=lower is better.}
\label{tab:thinkgeo_comparison_full_metrics}
\begin{tabular}{llcccc}
\toprule
\textbf{LLM Backbone} & \textbf{Framework} & \textbf{TSR (\%) ↑} & \textbf{FPA (\%) ↑} & \textbf{NTC ↓} & \textbf{NI ↓} \\
\midrule
\multirow{3}{*}{\textbf{Llama3-70B}} & ReAct & 7.2 & 6.0 & $\sim$25 & $\sim$18 \\
& Reflexion & 8.0 & 7.1 & $\sim$24 & $\sim$16 \\
& \textbf{Ours (CangLing-KnowFlow)} & \textbf{13.2} & \textbf{11.5} & \textbf{$\sim$18} & \textbf{$\sim$10} \\
\midrule
\multirow{3}{*}{\textbf{GPT-4}} & ReAct & 9.5 & 8.1 & $\sim$13 & $\sim$15 \\
& Reflexion & 10.5 & 9.2 & $\sim$1250 & $\sim$13 \\
& \textbf{Ours (CangLing-KnowFlow)} & \textbf{16.8} & \textbf{14.2} & \textbf{$\sim$9.5} & \textbf{$\sim$8} \\
\midrule
\multirow{3}{*}{\textbf{Claude 4 Sonnet}} & ReAct & 9.0 & 7.8 & $\sim$1350 & $\sim$16 \\
& Reflexion & 10.0 & 8.9 & $\sim$13 & $\sim$14 \\
& \textbf{Ours (CangLing-KnowFlow)} & \textbf{20.3} & \textbf{17.5} & \textbf{$\sim$9} & \textbf{$\sim$7} \\
\bottomrule
\end{tabular}
\end{table*}

To verify the cross-domain generalizability of our framework, we extended our evaluation to the public ThinkGeo benchmark \cite{shabbir2025thinkgeo}. 
We consider this experiment a pivotal validation step. 
It serves as a crucial test to determine whether the agent's capabilities are intrinsic to its architecture rather than merely artifacts of overfitting to specific internal tools and tasks. 
By subjecting the agent to ThinkGeo's distinct inventory of 14 tools and diverse tasks, we strictly evaluated its capacity to adopt its structured reasoning and dynamic adjustment mechanisms to an unseen case.

We benchmarked our framework against baselines using a selection of representative LLM backbones. 
The results, detailed in Table \ref{tab:thinkgeo_comparison_full_metrics}, reveal a compelling trend: even when operating within an unfamiliar toolset, CangLing-KnowFlow maintains a significant performance advantage. 
This demonstrates that our core, i.e., leveraging knowledge-guided structured workflows and dynamic adjustments, remains effective regardless of the specific tool environment, thereby confirming the architectural soundness and high transferability of our approach.

\subsection{Ablation Study}
To quantify the individual contributions of our framework's core components, we conducted an ablation study using the Claude 4 Sonnet backbone. 
We systematically disabled the Workflow Library (WL), Dynamic Adjustment (DA), and Learning Capability (LC). 
As shown in Table \ref{tab:ablation_study}, the results confirm that each module is indispensable to the system's overall performance.

\begin{table*}[!t]
\centering
\caption{Ablation study on the core components of our proposed method (using Claude 4 Sonnet as the backbone). We evaluate the impact of the Workflow Library (WL), Dynamic Adjustment (DA), and Learning Capability (LC). The best results are in \textbf{bold}.}
\label{tab:ablation_study}
\begin{tabular}{lcccc}
\toprule
\textbf{Configuration} & \textbf{w/o LC} & \textbf{w/o DA} & \textbf{w/o WL} & \textbf{Ours (Full Model)} \\
\midrule
\textit{Components} \\
\quad Workflow Library (WL) & \cmark & \cmark & \xmark & \cmark \\
\quad Dynamic Adjustment (DA) & \cmark & \xmark & \cmark & \cmark \\
\quad Learning Capability (LC) & \xmark & \cmark & \cmark & \cmark \\
\midrule
\textit{Performance Metrics} \\
Task Success Rate (\%) ↑ & 92.8 & 91.5 & 88.7 & \textbf{96.1} \\
First-Pass Accuracy (\%) ↑ & 73.6 & 75.9 & 67.1 & \textbf{79.2} \\
\midrule
\multicolumn{5}{l}{\textit{Planning Efficiency}} \\
\quad Number of Tool Calls ↓ & 7.9 & 8.6 & 13.2 & \textbf{6.8} \\
\quad Number of Interactions ↓ & 1.2 & 1.5 & 4.3 & \textbf{0.8} \\
\bottomrule
\end{tabular}%
\end{table*}

Specifically, the most substantial performance degradation occured upon the removal of the Workflow Library. 
In this configuration, the agent, stripped of its expert-guided procedural knowledge, reverted to a purely LLM-driven, unconstrained planning mode. 
This led to a dramatic decline in the TSR to 88.7\% and a sharp drop in FPA to 67.1\%. 
Notably, planning efficiency deteriorated significantly, as evidenced by by the surge in NTC to 13.2 and NI to 4.3. 
This outcome validates our central hypothesis, an expert-informed knowledge base is the foundational basis that prevents planning hallucination, constraining the agent to operational pathways that are both methodologically sound and computationally efficient.

Conversely, disabling the Dynamic Adjustment module provided critical insights into the framework's operational robustness. 
While the FPA remained relatively high at 75.9\%, reflecting the quality of the initial plans derived from the WL, the final TSR decreased to 91.5\%. 
This divergence between initial plan quality and final success highlights a crucial reality of real-world applications: even well-structured plans are vulnerable to failures stemming from data variability or environmental stochasticity. 
The DA module acts as the essential safety net, endowing the agent with the resilience required to navigate these unforeseen challenges and ensure task completion. 
Its absence renders the agent brittle, leaving it unable to recover from otherwise manageable execution errors.

Finally, the removal of the Learning Capability module resulted in a moderate yet discernible decline in accuracy (TSR of 92.8\%). 
The LC functions as a long-term optimization engine. 
By codifying validated \textit{ad-hoc} repairs into new workflow templates and attributing failures to heuristic rules, it continuously refines the agent's knowledge base. 
While its absence does not severely impair immediate problem-solving abilities, it restricts the evolutionary potential of the agent, preventing gains in efficiency and autonomy over time. 
This suggests that the utility of the LC module is underestimated in this short-term study; we anticipate its value would be significantly amplified in long-term implementations where the benefits of cumulative learning can fully accrue. 
This component is key to transitioning the agent from a static tool to an evolving, intelligent system.

In summary, the ablation study confirms that the superior performance of CangLing-KnowFlow stems not from any single feature, but from the orchestrated synergy of its components. 
The Workflow Library provides essential strategic guidance, the Dynamic Adjustment module ensures tactical robustness, and the Learning Capability enables long-term evolution.

\subsection{LLM's Intrinsic Capability}

A pivotal question in agentic systems is the interplay between the agent's architectural design and the intrinsic capabilities of its underlying LLM. 
Our experiments provide a clear answer: while a more capable LLM undoubtedly elevates performance, a well-designed framework like our CangLing-KnowFlow acts as a decisive force multiplier, often proving even more critical than the raw intelligence of the model itself.

\begin{table*}[!t]
\centering
\caption{Intrinsic capability evaluation of different large language models on core remote sensing agent tasks. These tests are conducted without any advanced agent framework to isolate the models' inherent abilities. The best results are in \textbf{bold}.}
\label{tab:llm_capability}
\begin{tabular}{lcccc}
\toprule
\textbf{LLM} & \textbf{\shortstack{Inst Acc. \\ (\%) ↑}} & \textbf{\shortstack{Tool Sel. F1 \\ (\%) ↑}} & \textbf{\shortstack{Param. Acc. \\ (\%) ↑}} & \textbf{\shortstack{Overall \\ Score ↑}} \\
\midrule
GPT-3.5-Turbo & 61.8 & 66.2 & 64.7 & 64.2 \\
Llama3-70B & 70.1 & 74.3 & 72.8 & 72.4 \\
Claude 3 Sonnet & 74.6 & 77.1 & 76.3 & 76.0 \\
Gemini 1.5 Pro & 80.7 & 83.6 & 81.4 & 81.9 \\
Claude 3 Opus & 83.4 & 85.9 & 84.7 & 84.7 \\
GPT-4 & 86.5 & 88.0 & 87.0 & 87.2 \\
GPT-4 Turbo & 87.8 & 89.3 & 88.5 & 88.5 \\
Claude 3.5 Sonnet & 89.2 & 90.6 & 89.8 & 89.9 \\
Claude 3.7 Sonnet & 90.6 & 91.9 & 91.3 & 91.3 \\
GPT-5 & 91.4 & 92.7 & 92.1 & 92.1 \\
\textbf{Claude 4 Sonnet} & \textbf{92.7} & \textbf{93.8} & \textbf{93.2} & \textbf{93.2} \\
\bottomrule
\end{tabular}
\end{table*}

As evidenced in Table \ref{tab:llm_capability}, there is a large positive correlation between an LLM's inherent abilities, such as instruction following (Inst Acc.), tool selection (Tool Sel. F1), and parameter accuracy (Param. Acc.), and its potential as an agent backbone. Top-tier models like Claude 4 Sonnet and GPT-5 demonstrate superior foundational skills. However, a shortsighted focus on the LLM alone would be misleading. The more profound insight emerges from cross-referencing these intrinsic capabilities with the end-to-end task performance in Table \ref{tab:performance_comparison}. For example, Llama3-70B \cite{grattafiori2024llama}, a model with respectable but not top-tier intrinsic scores, achieves a TSR of 83.6\% (simple) and 74.6\% (complex) when empowered by our CangLing-KnowFlow framework. It is notably better to that of a far more powerful model, GPT-4 \cite{achiam2023gpt}, when constrained to the ReAct framework (TSR of 82.9\% / 74.9\%). This demonstrates that the CangLing-KnowFlow architecture does not merely facilitate the LLM's existing abilities; it fundamentally enhances and guides them, enabling a moderately capable model to outperform a superior one that lacks proper guidance. 

\subsection{An Intuitive Comparison}

To provide a qualitative and intuitive illustration of CangLing-KnowFlow's practical advantages, we present a case study based on a representative RS task, as depicted in Figure \ref{fig:task_support}. The task involves building change detection and requires a multi-step workflow, including critical data preprocessing to ensure the accuracy of the final analysis. The divergent approaches taken by different agent frameworks are highly illustrative.

\begin{figure*}[!t]
\centering
\includegraphics[width=\linewidth]{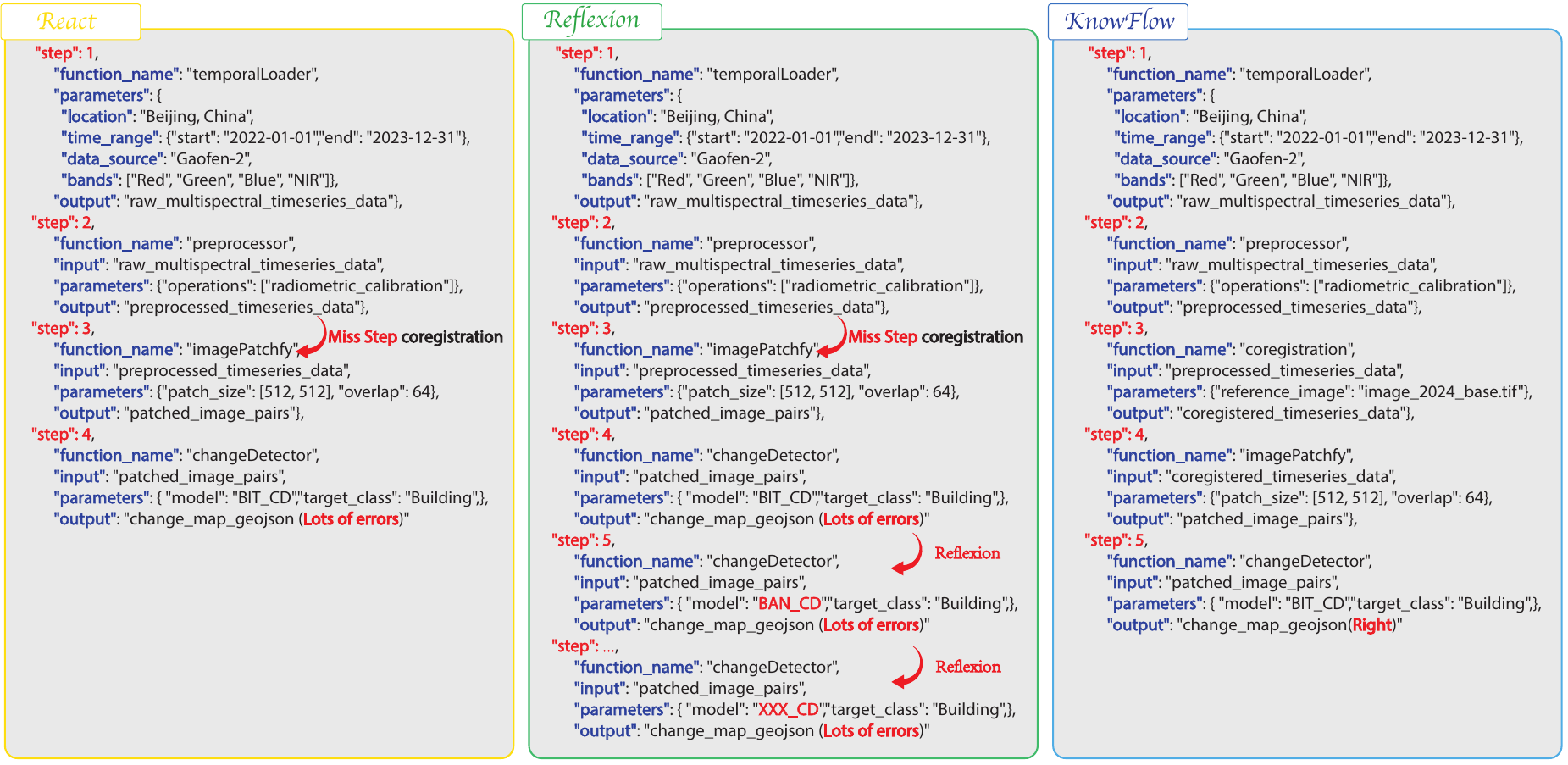}
\caption{Comparing the results in the visualization, ReAct missed Step 3 and obtained a poor result. Reflexion also missed Step 3 and tried to get a better result by changing the model. Our CangLing-KnowFlow got the correct tool arrangement.}
\label{fig:result}
\end{figure*}

We can first observe that the ReAct agent, operating on a step-by-step, reactive basis, exhibits a critical form of ``planning shortsighted''. Lacking a holistic understanding of the entire scientific workflow, it overlooks a crucial preprocessing step (Step 3 in the correct workflow). It proceeds directly to the core analysis, feeding improperly prepared data into the model. This fundamental process error inevitably leads to a scientifically unsound and incorrect final result.

As for the Reflexion agent, it demonstrates a marginal improvement. It successfully executes its initial, flawed plan and, upon observing a poor or nonsensical output, correctly identifies that a failure has occurred. However, its ``reflection'' is superficial. It attempts to correct the issue by altering local parameters or retrying the failed step, but it is incapable of diagnosing the root cause, which is a missing and fundamental step in the workflow itself. It is trapped within the confines of its flawed initial plan, unable to perform the necessary global restructuring.

By contrast, our CangLing-KnowFlow agent succeeds by its elaborate design. When a task begins, it queries its PKB and retrieves an expert-validated workflow template for this type of analysis. This template explicitly includes the mandatory preprocessing step. By adhering to this structured, scientifically vetted plan, CangLing-KnowFlow ensures that every action is performed in the correct sequence with the correct dependencies, guaranteeing the integrity of the process and the accuracy of the final result.

This case study clearly demonstrates how CangLing-KnowFlow, by integrating knowledge with workflow control, effectively overcomes the limitations inherent in less structured agent frameworks.

\section{Conclusion}\label{conclusion}

In this paper, we introduced CangLing-KnowFlow, a novel agent framework that synergistically fuses an expert-informed PKB, a robust Dynamic Adjustment mechanism, and an experience-driven Evolutionary Memory. 
It addresses several critical limitations of existing agentic systems directly by mitigating planning hallucinations and enhancing robustness against runtime uncertainties. 

Empirical validations across diverse benchmarks confirm the efficacy and robustness of our approach. On the domain-specific KnowFlow-Bench, the framework (powered by Claude 4 Sonnet) achieved peak Task Success Rates of 96.1\% and 95.1\% on simple and complex tasks, respectively, significantly outperforming ReAct baselines while reducing tool calls by approximately 38\%. 
We observed a similar trend on the ThinkGeo benchmark, where our method achieved more than double the baseline performance, increasing from 9.0\% to 20.3\% in unseen environments. 
Furthermore, ablation studies quantitatively identified the expert-informed Workflow Library as the foundational element for preventing planning hallucinations, with its removal causing the most significant degradation in both accuracy and efficiency.

By endowing agents with the cognitive capacities to reason, adapt, and learn within complex, dynamic environments, rather than merely executing static procedures, we establish a new paradigm for efficient, reliable, and scalable discovery.
Looking ahead, our research will advance this paradigm along several key frontiers. 
We will focus on expanding the PKB through semi-automated knowledge acquisition and engineering multi-agent collaborative workflows for highly complex scientific campaigns. 
More profoundly, we will direct CangLing-KnowFlow’s core principles toward accelerating mechanism-driven scientific discovery in new domains where procedural integrity is paramount. 
For example, a primary objective can be harnessing this framework to unravel the complex, interdisciplinary processes central to fields such as global sustainability and ecology, where RS science is poised to unlock fundamental new insights \cite{ghamisi2025geospatial}. 
This will position CangLing-KnowFlow not merely as an observation tool, but as a proactive partner in scientific inquiry.

\bibliographystyle{ieeetr}

\bibliography{reference}


\vfill

\end{document}